\documentclass[10pt,letterpaper,twocolumn]{article}
\usepackage[latin9]{inputenc}
\usepackage{float}
\usepackage{amsmath}
\usepackage{amssymb}
\usepackage{graphicx}
\usepackage[unicode=true,pdfusetitle,
 bookmarks=false,
 breaklinks=true,pdfborder={0 0 1},backref=section,colorlinks=false]
 {hyperref}
\hypersetup{
 pagebackref=true,letterpaper=true,colorlinks}

\makeatletter

\pdfpageheight\paperheight
\pdfpagewidth\paperwidth

\floatstyle{ruled}
\newfloat{algorithm}{tbp}{loa}
\providecommand{\algorithmname}{Algorithm}
\floatname{algorithm}{\protect\algorithmname}

\usepackage[T1]{fontenc}

\usepackage{cvpr}\usepackage{times}\usepackage{epsfig}
\graphicspath{{.},{./images/}}

\usepackage{caption}\usepackage{subcaption}%\usepackage{algpseudocode}
\usepackage{algorithm}\usepackage{algorithmic}

\newcommand{\mb}{\mathbf}

\newcommand{\denselist}{\itemsep 0pt\parsep=1pt\partopsep 0pt}
\newcommand{\bitem}{\begin{itemize}\denselist}
\newcommand{\eitem}{\end{itemize}}
\newcommand{\benum}{\begin{enumerate}\denselist}
\newcommand{\eenum}{\end{enumerate}}

\newcommand{\R}{\mathbb{R}}

\def\cl{{\cal L}}
\def\ch{{\cal H}}

\def\real{\mathbb{R}}

\def\st{\mathrm{s.t.}}
\newcommand{\trace}{\mathrm{tr}}
\def\tn{\textnormal}

 \cvprfinalcopy % *** Uncomment this line for the final submission

\def\cvprPaperID{680} % *** Enter the CVPR Paper ID here

\ifcvprfinal\fi

\makeatother

\begin{document}

\title{Graph Matching with Anchor Nodes: A Learning Approach}

\author{Nan Hu \hspace{20pt} Raif M. Rustamov \hspace{20pt} Leonidas Guibas\\
 Stanford University\\
 Stanford, CA, USA\\
 \texttt{\small nanhu@stanford.edu, rustamov@stanford.edu, guibas@cs.stanford.edu}}
\maketitle
\begin{abstract}

In this paper, we consider the weighted graph matching problem with
partially disclosed correspondences between a number of anchor nodes. Our
construction exploits recently introduced node signatures based
on graph Laplacians, namely the Laplacian family signature (LFS) on the
nodes, and the pairwise heat kernel map on the edges. In this paper, without
assuming an explicit form of parametric dependence nor a distance
metric between node signatures, we formulate an optimization problem
which incorporates the knowledge of anchor nodes. Solving this problem
gives us an optimized proximity measure specific to the graphs under
consideration. Using this as a first order compatibility term, we
then set up an integer quadratic program (IQP) to solve for a near
optimal graph matching. Our experiments demonstrate the superior performance
of our approach on randomly generated graphs and on two widely-used image
sequences, when compared with other existing signature and adjacency
matrix based graph matching methods.

\end{abstract}
%%%%%%%%% BODY TEXT

\section{Introduction}

The exact and approximate graph matching problem is of great interest
in computer vision due to its numerous applications in areas such
as 2D and 3D image registration, object recognition and biomedical
identification, and object
tracking in video sequences. An important variant of the problem is
the semi-supervised setting where a small proportion of correct node
correspondences between the graphs are known. Such correspondences can
be based on additional information provided for only a few nodes, human
judgement or prior knowledge, etc. Algorithms that can take
advantage of this information to infer the correspondences for the
rest of the graph nodes are highly desirable.

While there has been a recent effort in applying machine learning
concepts to the graph matching problem in computer vision \cite{caetano2009,Leordeanu2012},
these works are based on the assumption that a training set consisting
of pairs of graphs with fully correct correspondences given, and that the
training set is representative enough of testing graphs, so that learning
done with training graphs can be usefully transferred to testing graphs. This
problem setup, however, is different from our setting where
we only have two graphs with partially known correspondences; in a
sense, these known correspondences constitute our ``training data''.
In addition, this amounts to a much smaller and more restricted amount
of training data, making the problem challenging.

In this work we provide a method for effectively incorporating known
correspondences into the commonly used integer quadratic program formulation
of graph matching. Specifically, our main contribution is to devise
a new first order compatibility term between two nodes of different
graphs. Our method uses the recently proposed one-parameter family
of node signatures called Laplacian Family Signatures (LFS) \cite{hu2012},
which provide a feature vector (signature) for each node based solely
on the node's structural position within the graph. In contrast to
\cite{hu2012}, we do not assume an explicit form of parametric dependence
for generating these signatures, but leave it in an unspecified generic
form. Since graph matching is performed using the dissimilarity/distance
between these signatures, we derive the distance between the generic
signatures. As a result of this manipulation, we find that the entire
process of computing and comparing these generic signatures can be
encoded into a single proximity matrix. We then introduce an algorithm
to learn this proximity matrix from the knowledge of provided correct
correspondences. This is done by requiring anchor nodes in one graph to correspond near
to their known partners in the other and to be far from non-correspondences, which
can be set up as a max-margin problem \cite{Xing2002}.

This method was chosen due to a number of benefits. First, our max-margin
formulation makes an effective use of the scarce training data:
even a small number of known correspondences (two anchor correspondences
are used in all of our experiments) leads to a large number of constraints
on the proximity matrix. Second, our formulation results in learning
a proximity matrix that is relatively small (tens by tens in the examples
shown) which allows us to reliably learn it without over-fitting. Third,
our max-margin problem, can be solved using column generation \cite{Lubbecke2005},
which results in an efficient algorithm that scales well with the
increasing size of the graphs and number of constraints.

\paragraph*{Notation and Paper Organization }

Let $G=(V,E)$ and $G'=(V',E')$ be two undirected weighted graphs.
Our goal is to find an approximate matching between these two graphs
based solely on their structural properties (e.g.~no externally provided
attributes are available for nodes). We assume that a partial correspondence
between graphs is given. Namely, let $U\subset V$ and $U'\subset V'$
be the subsets of nodes that are known to be in correspondence; we
will refer to these as anchor node sets for $G$ and $G'$ respectively.

We follow the commonly used integer quadratic program (IQP) formulation
for the graph matching problem based on two kinds of compatibility
terms. The first order compatibility $d(i,a)$ encodes similarity
of a node $i\in V$ in graph $G$ to a node $a\in V'$ in graph $G'$.
For pairs of nodes $i,j\in V$ and $a,b\in V'$, the second order
compatibility $d(i,j,a,b)$ measures the compatibility of matching
the node pair $(i,j)$ to node pair $(a,b)$.
%node $i$ to $a$ with matching node $j$ to $b$.

Our main goal in this paper is to incorporate the knowledge of the
anchor correspondences into the first order compatibility term which
will be expressed as follows:
\[
d(i,a)=c_{B}d_{B}(i,a)+c_{\text{ap}}d_{\text{ap}}(i,a),
\]
where $c_{B}$ and $c_{ap}$ are some weights. The first term $d_{B}(i,a)$
is based on our formulation of LFS proximity matrix and is presented
in Section \ref{ssect:lfs}. The construction of the second term $d_{\text{ap}}(i,a)$
which involves the heat kernel is explained in Section \ref{ssect:hdd}.
The matching scheme incorporating both the first order and second
order compatibility functions is presented in Section \ref{sect:matchingscheme}.
In Section \ref{sect:exp}, we present experiments using our algorithm
on three common datasets.

\section{Related Work}

\label{ssect:relatedwork}

Our family of signatures are closely related to node-based signatures
on graphs, different forms of which has already been considered,
e.g. %Joilli et. al.
\cite{eshera1984,shok2001,gori2005,Wong2006,jouili2009}. Recently,
Sun et. al \cite{sog-hks-09} proposed the heat kernel signature (HKS)
for application of shape matching in geometry processing. Their signature is based on the
simulated heat diffusion process on manifold. Aubry et. al. \cite{Aubry2011}
later proposed a signature of similar structure based on quantum processes
on graph. Both have been shown in \cite{hu2012} to be special instances
of LFS.

Different forms of spectral matrices have already been considered
in matching. Among the pioneering work is Umeyama's \cite{umeyama88}
weighted graph matching algorithm from a decomposition of adjacency
matrices. %His algorithm restrict the graphs are at least near isomorphism. Under this assumption, he derived a (closed-form) solution from the eigendecomposition of adjacency matrices that optimize the objective function.
His method was later generalized to graphs of different sizes \cite{luo2001,zhao2007}.
Robles-Kelly et. al. \cite{robles-kelly2002} used the steady state
of the Markov transition matrix to order the nodes and match using edit
distances.% from a symmetrically-normalized adjacency matrix together with the edge connectivity constraint to order the nodes of the graph into a string for a matching using edit distance.
%Matching is computed as the minimizer of a graph edit distance with the edit transition probability defined to be a combination of node similarity and edge connectivity.
%In their 2005 paper
Later in \cite{robles-kelly2005}, the same authors proposed to use
the leading eigenvector of the adjacency matrix to serialize the graph
nodes for matching. % and computed the matching using a similar edit distance algorithm.
Qiu et. al \cite{qiu2006} considered using the Fiedler vector together
with the proximity to the perimeter of the graph to partition the
graph into disconnected components for a hierarchical matching. %A hierarchical matching is applied to each level of partitions using graph edit distance. Their method, however, works only on planar graphs as the computation of the perimeter of a graph in general is a NP-complete problem.
Cho et. al. \cite{cho2010} constructed a reweighed random walk similar
to personalized PageRank on the association graph with the addition
of an absorbing node, and used the quasi-stationary distribution to
find a matching. %They computed the quasi-stationary distribution of the random walk and discretize the continuous solution to find a matching. Recently, there is an interest of incorporating quantum computing concepts into graph matching, e.g.
Emms et. al. \cite{emms2009} %built an auxiliary graph from the two graphs by adding auxiliary vertices connecting every pair of vertices from each graph.
simulated a quantum walk on the auxiliary graph % and the interference amplitude (equivalently probability) of particles on each auxiliary nodes are characterized over time.
%Experiment showed that for a heuristically selected time the interference amplitude is small for true assignment and large for false assignment. Hence, t
%They
and used the particle probability of each auxiliary node as the cost
of assignment for a bipartite matching.%, with two rounds of Hungarian algorithm for a matching.
%is applied to find the min-cost assignment with the consideration of edge consistency.
%On the down side, however, the complexity of their algorithm is $O(n^6)$.

In a broader sense, other relaxation-based
matching algorithms are also related to our work.
%Most of these algorithms were intended to solve a relaxation of the original integer quadratic problem (IQP).
Gold and Rangarajan \cite{gold1996} proposed the well-known \textit{Graduated
Assignment Algorithm}. %The idea of their (discrete annealing) algorithm is to start with a solution that is not guaranteed to satisfy all constraints, and iteratively move toward a consistent solution that is compatible with more stringent constraints.
%The algorithm, however, does not necessarily converge to the optimal solution.
van Wyk et. al. \cite{vanwyk2004} designed a projection onto convex
set (POCS) based algorithm to solve IQP. % by successively project the relaxation solution onto the convex constrain set.
Schellewald et. al. \cite{schell2005} constructed a semi-definite
programming relaxation of the IQP. Leordeanu et. al. \cite{leord2005}
proposed a spectral method to solve a relaxed IQP by only considering
linear inequality constraints at discretization. % where they drop off the linear inequality constraint during relaxation and only incorporate it at the discretization step.
The idea was further extended by Cour et. al. \cite{cour2006}, where
they added an affine constraint during relaxation. Zaslavskiy et.
al. \cite{zas2009} approached the IQP from the point of a relaxation
of the original least-square problem to a convex and concave optimization
problem on the set of doubly stochastic matrices. Leordeanu et. al.
\cite{leord2009} proposed an integer projected fixed point (IPFP)
algorithm to %solve the quadratic assignment problem of graph matching by
iteratively search for a fixed point solution and then discretize
it into the matching domain.

\section{Anchor Based Compatibility}

\label{sect:descriptors}In this section we discuss the construction
and computation of two kinds of first order compatibility terms that
take advantage of known correspondences between anchor nodes.

\subsection{Generic Node Signatures and Their Comparison}

\label{ssect:lfs} We start by reviewing the concept of Laplacian
Family Signatures introduced in \cite{hu2012}. Consider one of the
graphs to be matched, say $G=(V,E)$. Let $w$ be the weights on edges,
i.e. $w:E\mapsto\mathbb{R}^{+}$. The graph Laplacian is defined as
$\cl=D-A$, where $A$ is the graph adjacency matrix with
\[
A_{ij}=\begin{cases}
w(i,j) & \text{if}~~(i,j)\in E\\
0 & \text{otherwise}
\end{cases}
\]
and $D$ is a diagonal matrix of total incident weights, i.e. $D_{ii}=\sum_{j}{A_{ij}}$.
$\cl$ has numerous nice properties \cite{biyi2007}, of which most relevant to us
is the symmetry and positive semi-definiteness.
This makes it possible to consider the eigen-decomposition of $\cl$; we denote
by $\{\lambda_{k},\phi_{k}\}_{k=1}^{|V|}$ the eigenpairs of graph
Laplacian matrix $\cl$ (eigenvalue and associated eigenvector).
We use the same notation with the prime symbol
added for the corresponding constructs of our second graph $G'$.

The Laplacian Family Signatures (LFS) for a node $u\in V$ is a one-parameter
family of structural node descriptors that is defined by
\begin{equation}
s_{u}(t)=\sum_{k}h(t;\lambda_{k})\phi_{k}(u)^{2}\label{eqn:lfs}
\end{equation}
where $h(t;\lambda_{k})$ is a real valued function. Special $h(t;\lambda_{k})$
of different forms will result in the heat kernel signature (HKS)
\cite{sog-hks-09} when $h(t;\lambda_{k})=\exp(-t\lambda_{k})$, the
wave kernel signature (WKS) \cite{Aubry2011} when
$h(t;\lambda_{k})=\exp(-\frac{(t-\log\lambda_{k})^{2}}{2\sigma^{2}})$,
or the wavelet signature if $h(t;\lambda_{k})$ admits some special behavior
as described in \cite{HAMMOND2011}.

These signatures describe a given node's structural relationship to
its neighborhood. For example, HKS has an interpretation in terms
of a simulated heat diffusion process: for each node, this signature
captures the amount of heat left at the node at various times (here
$t$) assuming that a unit amount is put on the node initially ($t=0$).
These signatures are naturally intrinsic, namely if two graphs are
isomorphic, then the signatures of corresponding nodes are the same;
the signatures are also stable under small perturbations \cite{hu2012}.

The above discussion suggests using these signatures as node attributes
to design first order compatibility terms --- for if the signatures of
two nodes from the two graphs are very different, then these vertices
are less likely to be in correspondence. However, such an approach
does not take into account the given anchor correspondences, because
the form of the function $h(t;\lambda_{k})$ is explicitly provided
beforehand.

To overcome this difficulty, in this paper, in contrast to \cite{hu2012},
we will not assume an explicit form for the function $h(t;\lambda_{k})$,
nor will we assume a specific form of dissimilarity measure when comparing the
LFS of two nodes. Instead, we assume that $h(t;\lambda_{k})$ is a
generic linear combination of some real-valued functions $\{b_{i}(t)\}_{i=1}^{N_{b}}$,
given as
\begin{equation}
h(t;\lambda_{k})=\sum_{i=1}^{N_{b}}a_{ki}b_{i}(t),\label{eqn:basis}
\end{equation}
where $\{a_{ki}\}$ are some real coefficients. We assume a similar
expression for the second graph $G'$ with possibly a different set
of coefficients $\{a_{ki}'\}$. Let $\langle\cdot,\cdot\rangle$ be
an arbitrary inner product of real-valued functions. Assuming that
LFS comparison employs this dot product, the dissimilarity between
two nodes $u\in V,v\in V'$ can be expressed as
\[
\begin{array}{ll}
d^{2}\left(s_{u}(t),s'_{v}(t)\right) & =\langle s_{u}(t)-s'_{v}(t),s_{u}(t)-s'_{v}(t)\rangle\end{array}
\]
Substituting (\ref{eqn:basis}) to (\ref{eqn:lfs}), we have
\[
s_{u}(t)=\sum_{k=1}^{K}\phi_{k}^{2}(u)\sum_{i=1}^{N_{b}}a_{ki}b_{i}(t),
\]
\[
s'_{v}(t)=\sum_{k=1}^{K'}{\phi'}_{k}^{2}(v)\sum_{i=1}^{N_{b}}a'_{ki}b_{i}(t).
\]
Denote $A=\left[a_{ki}\right]\in\R^{K\times N_{b}}$, $\theta_{u}=\left[\begin{array}{l}
\phi_{1}^{2}(u)\\
\vdots\\
\phi_{K}^{2}(u)
\end{array}\right]\in\R^{K}$, $A'=\left[a'_{ki}\right]\in\R^{K'\times N_{b}}$, $\theta'_{v}=\left[\begin{array}{l}
{\phi'}_{1}^{2}(v)\\
\vdots\\
{\phi'}_{K'}^{2}(v)
\end{array}\right]\in\R^{K'}$. Let $b(t)=\left[\begin{array}{l}
b_{1}(t)\\
\vdots\\
b_{N_{b}}(t)
\end{array}\right]\in\R^{N_{b}}$ and $C_{ij}=\langle b_{i}(t),b_{j}(t)\rangle$. Now after denoting
$C=\left[C_{ij}\right]\in\R^{N_{b}\times N_{b}}$, we obtain

\[
\begin{array}{l}
d^{2}\left(s_{u}(t),s'_{v}(t)\right)\\
=\langle\theta_{u}^{\top}Ab(t)-{\theta'}_{v}^{\top}A'b(t),\theta_{u}^{\top}Ab(t)-{\theta'}_{v}^{\top}A'b(t)\rangle\\
=\left[\begin{array}{l}
\theta_{u}\\
\theta'_{v}
\end{array}\right]^{\top}\underbrace{\left[\begin{array}{l}
A\\
-A'
\end{array}\right]C\left[\begin{array}{l}
A\\
-A'
\end{array}\right]^{\top}}_{B\in\R^{(K+K')\times(K+K')}}\underbrace{\left[\begin{array}{l}
\theta_{u}\\
\theta'_{v}
\end{array}\right]}_{w_{uv}\in\R^{(K+K')}}\\
=w_{uv}^{\top}Bw_{uv}
\end{array}
\]

This formulation holds for any inner-product based dissimilarity metric.
The number of basis functions, although assumed finite above can be
easily extended to infinite, i.e. the formulation is still valid as
$N_{b}\rightarrow\infty$ and the basis per se is also arbitrary.
The only restriction, as a result of the positive semi-definiteness
of $C$, is $B\succeq0$.

The above discussion gives the general expression that we will use
as a part of our first-order compatibility measure. Namely, we set
$d_{B}(i,a)=\sqrt{w_{ia}^{\top}Bw_{ia}}$ for any two nodes $i\in V,a\in V'$.
This representation is especially useful since it avoids determining
the intermediate matrices $C$, $A$ and $A'$ explicitly, but allows
us to learn directly the proximity matrix $B$ which is a small matrix
(tens by tens in our experiments).

\subsubsection{Learning the Proximity Matrix}

Here we explain how to learn the proximity matrix $B$ from the knowledge
of anchor nodes. A good proximity matrix $B$ should move closer node
pairs that correspond, and move away nodes that are non-matches. If
we let $U\subset V$ be the set of anchor nodes in graph $G$, and
$U'\subset V'$ be their known correspondences in graph $G'$, then
we want $d_{B}^{2}(i,a)=w_{ia}^{\top}Bw_{ia}$ to be small if $i$
and $a$ are in correspondence, and to be large otherwise. One way
of achieving this is to formulate the problem as a max-margin
problem similar to SVM.

%Recall $G=(V,E),G'=(V',E')$ be the two graphs, and $U\subset V$ be the set of anchor nodes, $U'\subset V'$ be the known correspondences of $U$, and $d\left(s_i(t),s'_{j}(t)\right) = w_{ij}^\top Bw_{ij}$. Then the problem could be written as

%\[\begin{array}{ll}
%\Eps & = \sum_{u\in U}\frac{d^2\left(s_u(t),s'_{P(u)}(t)\right)}{\sum_{v\in V'\backslash P(u)}d^2\left(s_u(t),s'_{v}(t)\right)+\sum_{u'\in V\backslash u}d^2\left(s_{u'}(t),s'_{P(u)}(t)\right)} \\
%& = \sum_{u\in U}\frac{w_{u,P(u)}Bw_{u,P(u)}}{\sum_{v\in V'\backslash P(u)}w_{uv}Bw_{uv}+\sum_{u'\in V\backslash u}w_{u',P(u)}Bw_{u',P(u)}}
%\end{array}
%\]
%For each term, both the numerator and denominator are linear in $B$. We could then try to minimize $\Eps$ to find $B$.

\[
\begin{array}{ll}
\max & \gamma\\
\st & w_{ib}^{\top}Bw_{ib}-w_{ia}^{\top}Bw_{ia}\geq\gamma,~\forall i,~\forall b\neq a\\
 & w_{ja}^{\top}Bw_{ja}-w_{ia}^{\top}Bw_{ia}\geq\gamma,~\forall a,~\forall j\neq i\\
 & B\succeq0\\
 & \|B\|_{F}\leq1\,.
\end{array}
\]

This can be easily verified to be equivalent to

\[
\begin{array}{ll}
\min & \frac{1}{2}\|B\|_{F}^{2}\\
\st & \trace((w_{ib}w_{ib}^{\top}-w_{ia}w_{ia}^{\top})B)\geq1,~\forall i,~\forall b\neq a\\
 & \trace((w_{ja}w_{ja}^{\top}-w_{ia}w_{ia}^{\top})B)\geq1,~\forall a,~\forall j\neq i\\
 & B\succeq0\,.
\end{array}
\]

For large graphs, however, the problem could be very possibly infeasible.
Therefore, we allow some violation in the training set and introduce
slack variables.

\[
\begin{array}{ll}
\min & \frac{1}{2}\|B\|_{F}^{2}+C\frac{1}{n}\sum_{i=1}^{n}\xi_{i}\\
\st & \trace((w_{ik}w_{ik}^{\top}-w_{ij}w_{ij}^{\top})B)\geq1-\xi_{i},~\forall i,~\forall k\neq j\\
 & \trace((w_{lj}w_{lj}^{\top}-w_{ij}w_{ij}^{\top})B)\geq1-\xi_{i},~\forall j,~\forall l\neq i\\
 & B\succeq0\\
 & \xi_{i}\geq0\,,
\end{array}
\]
where $n=|U|$ is the number of anchor nodes.

One drawback of this formulation is that we put uniform weights on
slack variables. However, intuitively for a violation of the margin
constraint, we would rather to have the violated nodes to be near
to the correct matches within the graph, namely we want to put a non-uniform
scale on the slack variables to penalize more severely for nodes that
are farther from the correct matches. Therefore, we introduce a loss-function
$\Omega(k,j)$ to re-scale the slack variables. In our graph matching
setting, $\Omega(k,j)$ could be the shortest distance over the graph,
or the heat kernel as described in Section \ref{ssect:hdd} (we used
heat kernel in our experiments as it has been shown to be more robust
than the adjacency matrix \cite{hu2012} and, hence, shortest distance).
Now the problem becomes

\[
\begin{array}{ll}
\min & \frac{1}{2}\|B\|_{F}^{2}+C\frac{1}{n}\sum_{i=1}^{n}\xi_{i}\\
\st & \trace((w_{ib}w_{ib}^{\top}-w_{ia}w_{ia}^{\top})B)\geq1-\frac{\xi_{i}}{\Omega'(b,a)},~\forall i,~\forall b\neq a\\
 & \trace((w_{ja}w_{ja}^{\top}-w_{ia}w_{ia}^{\top})B)\geq1-\frac{\xi_{i}}{\Omega(j,i)},~\forall a,~\forall j\neq i\\
 & B\succeq0\\
 & \xi_{i}\geq0\,.
\end{array}
\]

Let $(\cdot)_{\tn{vec}}$ be the vector form of a matrix, and $\mb{b}=B_{\tn{vec}}$
and $\psi_{ik}=(w_{ik}w_{ik}^{\top}-w_{ij}w_{ij}^{\top})_{\tn{vec}}$.
The above problem could be solved by first relaxing the semi-definite
constraint and then projecting the solution to the semi-definite cone.
The relaxed problem is a quadratic programming problem

\[
\begin{array}{ll}
\min & \frac{1}{2}\|\mb{b}\|^{2}+C\frac{1}{n}\sum_{i=1}^{n}\xi_{i}\\
\st & \psi_{ib}^{\top}\mb{b}\geq1-\frac{\xi_{i}}{\Omega'(b,a)},~\forall i,~\forall b\neq a\\
 & \psi_{ja}^{\top}\mb{b}\geq1-\frac{\xi_{i}}{\Omega(j,i)},~\forall a,~\forall j\neq i\\
 & \xi_{i}\geq0\,.
\end{array}
\]

The dual of it is

\[
\begin{array}{ll}
\max & -\frac{1}{2}\sum_{ib}\sum_{ja}\alpha_{ib}\alpha_{ja}\psi_{ib}^{\top}\psi_{ja}+\sum\alpha_{ib}\\
\st & \alpha_{ib}\geq0\\
 & \sum_{i}\left(\frac{\alpha_{ib}}{\Omega'(b,a)}+\frac{\alpha_{ja}}{\Omega(j,i)}\right)\leq\frac{C}{n}\,.
\end{array}
\]

As the number of constraints in this problem is of $O(|U|(|V|+|V'|))$,
it becomes impossible to solve when the size of the graphs is very
large. One technique that could be used to lower the computational
cost is column generation \cite{Lubbecke2005}. The key idea of this
iterative algorithm is that although the number of constraints is
large, only a small portion of them will be nonzero at the solution.
Therefore, only this small subset of constraints are necessary to
the solution. To find this subset, the algorithm iteratively adds
one constraint per training sample that violated the constraint the
most until all constraints are satisfied. In addition, after each
iteration, we need to project $B$ back to semi-definite cone to restrain
$B\succeq0$. The pseudocode of the algorithm is omitted here for the sake of saving space. %The algorithm is summarized in Algorithm \ref{alg:mlearn}.

\subsection{Heat Kernel with Anchor Nodes}

\label{ssect:hdd} In this subsection we introduce the second term
appearing in our first order compatibility measure. This term is based
on the heat diffusion process on graph $G$. Specifically, consider
the graph heat kernel $k_{t}(u,v)$, which measures the amount of
heat transferred from node $u$ to node $v$ after time $t$, assuming
a unit amount was placed at $u$ in the beginning ($t=0$). The heat
kernel has the following representation in terms of the eigen-decomposition
of the graph Laplacian:
\[
k_{t}(u,v)=\sum_{k}\exp(-t\lambda_{k})\phi_{k}(u)\phi_{k}(v).
\]
We use the heat kernel value of anchor nodes at a given node as another
first order constraint. Namely, for any node $v$ of graph $G$ define
the heat kernel distance to anchor nodes as
\[
d_{\text{ap}}^{\ch}(v)=\sum_{u\in U}k_{t}(u,v),
\]
where $U$ is the set of anchor nodes of $G$. The same construction
using the anchor nodes $U'$ of our second graph $G'$ provides the
quantity $d_{\text{ap}}^{'\ch}(\cdot)$ for each node of $G'$.

For two nodes $i$ and $a$ in graphs $G$ and $G'$, we define the
second portion of our first order compatibility measure by
\[
d_{\text{ap}}(i,a)=\left|d_{\text{ap}}^{\ch}(i)-d_{\text{ap}}^{'\ch}(a)\right|.
\]
This quantity is a plausible measure of dissimilarity between nodes
because the anchor nodes $U$ and $U'$ are known to be in correspondence.
Moreover, the heat kernel is naturally intrinsic (if two graphs are
isomorphic, their corresponding heat kernels are the same) and it
is stable under small perturbations \cite{hu2012}.

\section{Matching Scheme}

\label{sect:matchingscheme}

Here we formulate the graph matching as an integer quadratic program
(IQP). Let $G=(V,E)$ and $G'=(V',E')$ be the two graphs, $U$ and
$U'$ be the anchor node set for $G$ and $G'$ respectively. For
any two nodes $i\in V\backslash U$ and $a\in V'\backslash U'$, let
$d_{B}^{2}(i,a)=w_{ia}^{\top}Bw_{ia}$ be the learned proximity, and
let $d_{\text{ap}}(i,a)=\left|d_{\text{ap}}^{\ch}(i)-d_{\text{ap}}^{'\ch}(a)\right|$.
Our first order compatibility term is
\[
d(i,a)=c_{B}d_{B}(i,a)+c_{\text{ap}}d_{\text{ap}}(i,a).
\]

Now we need a second order compatibility term, for which we will use
the heat kernel as done in \cite{hu2012}. For any two nodes $i,j\in V$
and $a,b\in V'$, the pairwise heat kernel distance is defined as
\[
d_{k}(i,j,a,b)=\left|k_{t}(i,j)-k'_{t}(a,b)\right|,
\]
and this gives a measure of how compatible matching nodes $i$ and
$a$ is with matching $j$ and $b$. As has been discussed in \cite{hu2012},
the heat kernel can be thought of as a noise tolerant approximation
of the adjacency matrix, and is stable under small perturbations.
Thus, our second order proximity term can be thought as a generalization
of the commonly used adjacency-based second order term.

Combining all this information, we construct a compatibility matrix
$W\in\real^{(|V|-|U|)(|V'|-|U'|)\times(|V|-|U|)(|V'|-|U'|)}$,
\[
W_{ia,jb}=\begin{cases}
d_{k}(i,j,a,b), & i\neq j,a\neq b\\
d(i,a), & i=j,a=b
\end{cases}
\]
Let $\mb{X}\in\{0,1\}^{(|V|-|U|)(|V'|-|U'|)}$ be the one-to-one mapping
matrix, and $\mb{x}\in\{0,1\}^{(|V|-|U|)(|V'|-|U'|)}$ be the vector
form of $\mb{X}$. The IQP can be written as
\[
\begin{aligned} & ~\mb{x}^{*}=\arg\max(\mb{x}^{\top}W\mb{x})\\
\text{s.t.}~ & ~\mb{x}\in\{0,1\}^{(|V|-|U|)(|V'|-|U'|)}\\
 & \forall i~\sum_{a\in V'\backslash U'}\mb{x}_{ia}\leq1,~\forall a~\sum_{i\in V\backslash U}\mb{x}_{ia}\leq1\,.
\end{aligned}
\]

As is well-known this problem is NP-complete and there is rich literature
on approximation algorithms for this problem. Comparison of the performance
of different IQP approximation solvers is outside the scope of our
paper. In our experiments, we selected a recently proposed algorithm,
reweighed random walk matching (RRWM) \cite{cho2010}. The main reason
we chose this algorithm is its superior performance when compared
with other state-of-the-art approximation solvers, including SM \cite{leord2005},
SMAC \cite{cour2006}, HGM \cite{zass2008}, IPFP \cite{leord2009},
GAGM \cite{gold1996}, SPGM \cite{vanwyk2004}. In consideration of
space, we omit the introduction of their algorithm here and leave
the details to the original paper \cite{cho2010}.

\section{Experiments}

\label{sect:exp}

We tested our approach on three different datasets: 1) synthetically
generated random graphs; 2) CMU Hotel sequence for large baseline
matching; and 3) pose house sequence from \cite{mcauley2010} for
large rotation angle matching.

\subsection{Synthetic Random Graphs}

\label{ssect:randgraph} In this section, following the experimental
protocol of \cite{cho2010}, we synthetically generated random graphs
and performed a comparative study. For a pair of graph $G_{1}$ and
$G_{2}$, they share $n_{\text{in}}$ common nodes and $n_{\text{out}}^{(1)}$
and $n_{\text{out}}^{(2)}$ outlier nodes. Edges are constructed with
a density $\rho$ and weights are randomly distributed in $[0,1]$.
Perturbation is done with added random Gaussian noise ${\cal N}(0,\sigma^{2})$.

\begin{figure*}[ht]
\centering \begin{subfigure}{.32\linewidth} \includegraphics[width=0.96\linewidth]{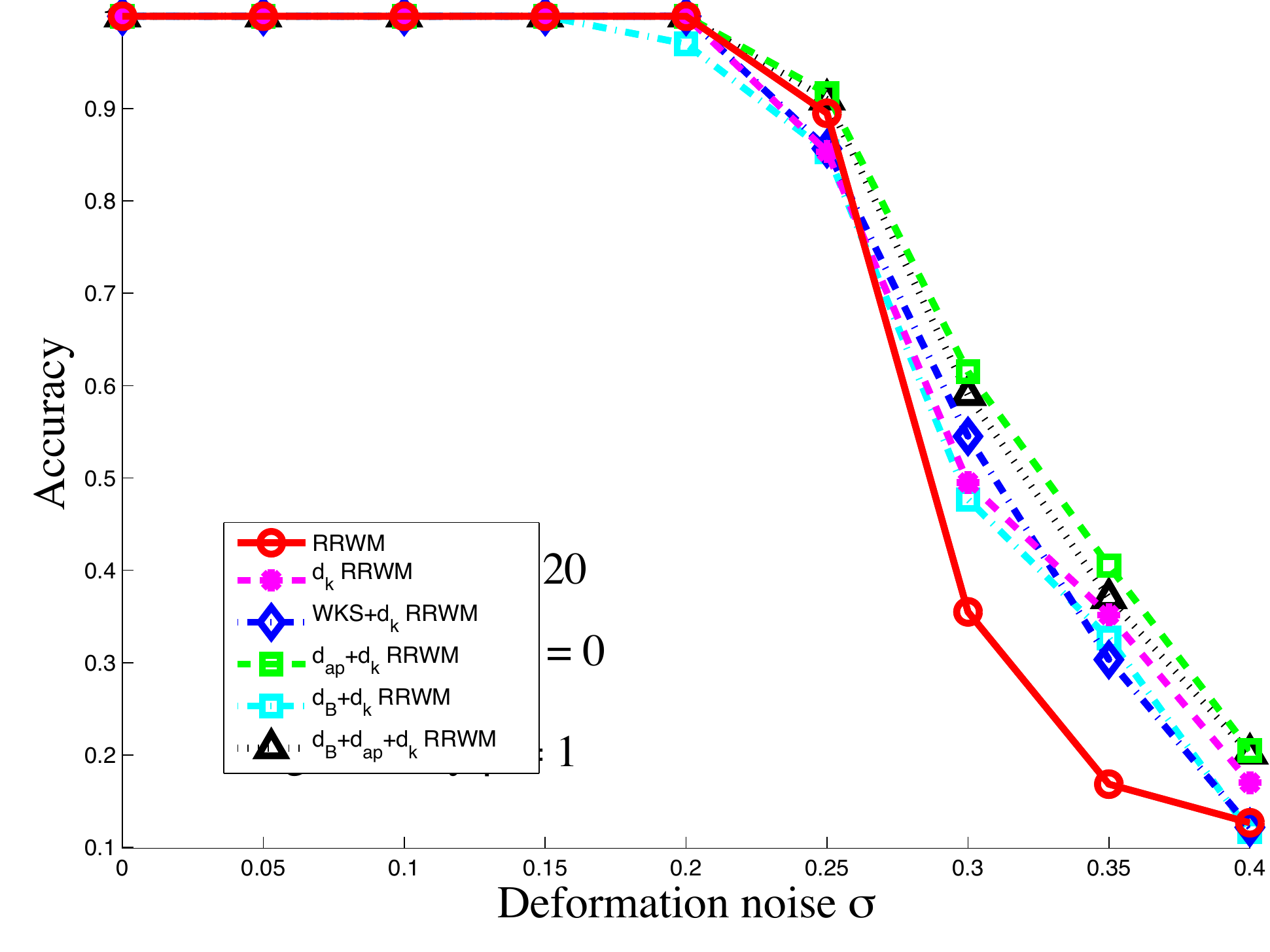}
\caption{}

\end{subfigure} \begin{subfigure}{.32\linewidth} \includegraphics[width=0.96\linewidth]{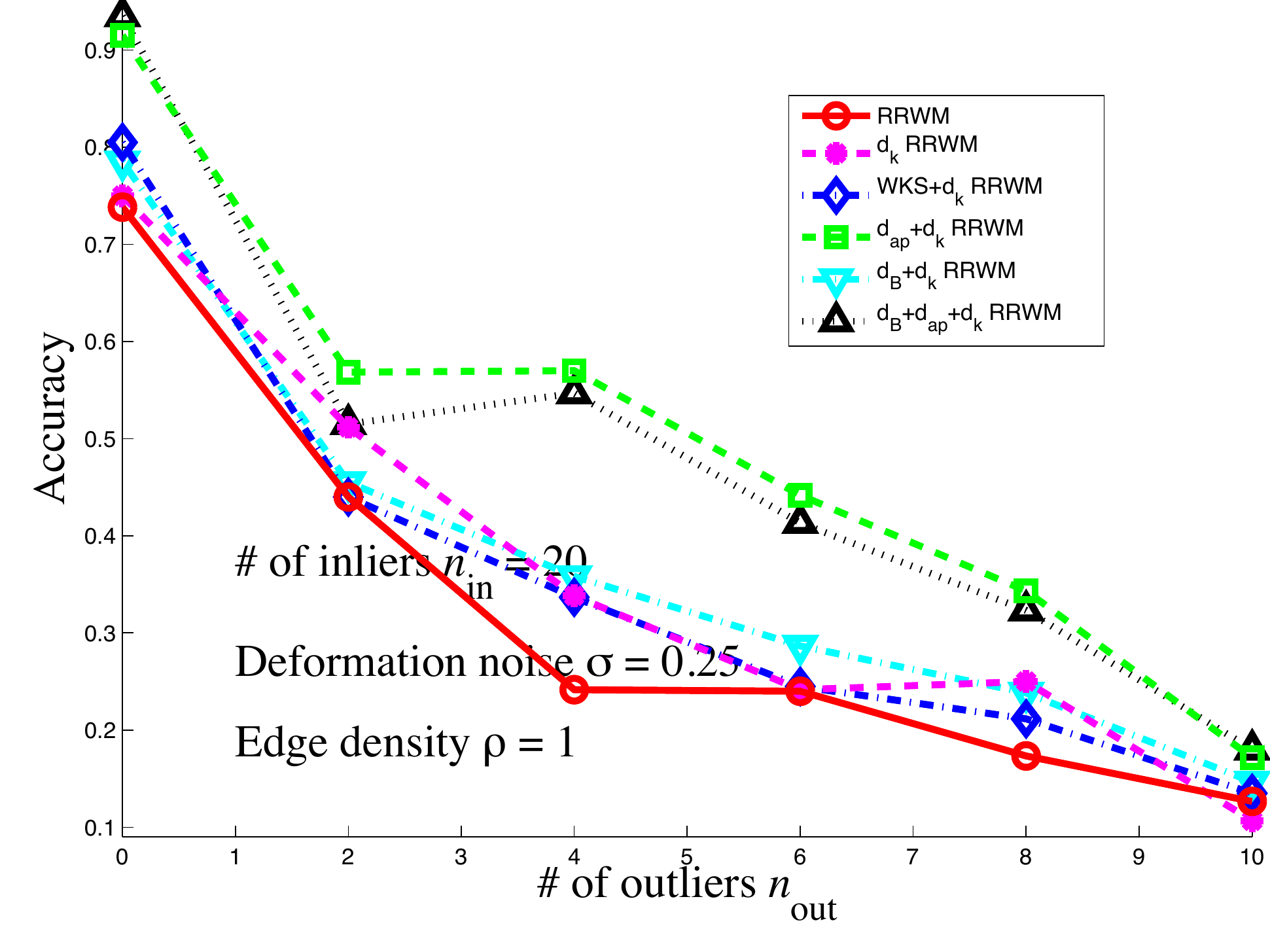}
\caption{}

\end{subfigure} \begin{subfigure}{.32\linewidth} \includegraphics[width=0.96\linewidth]{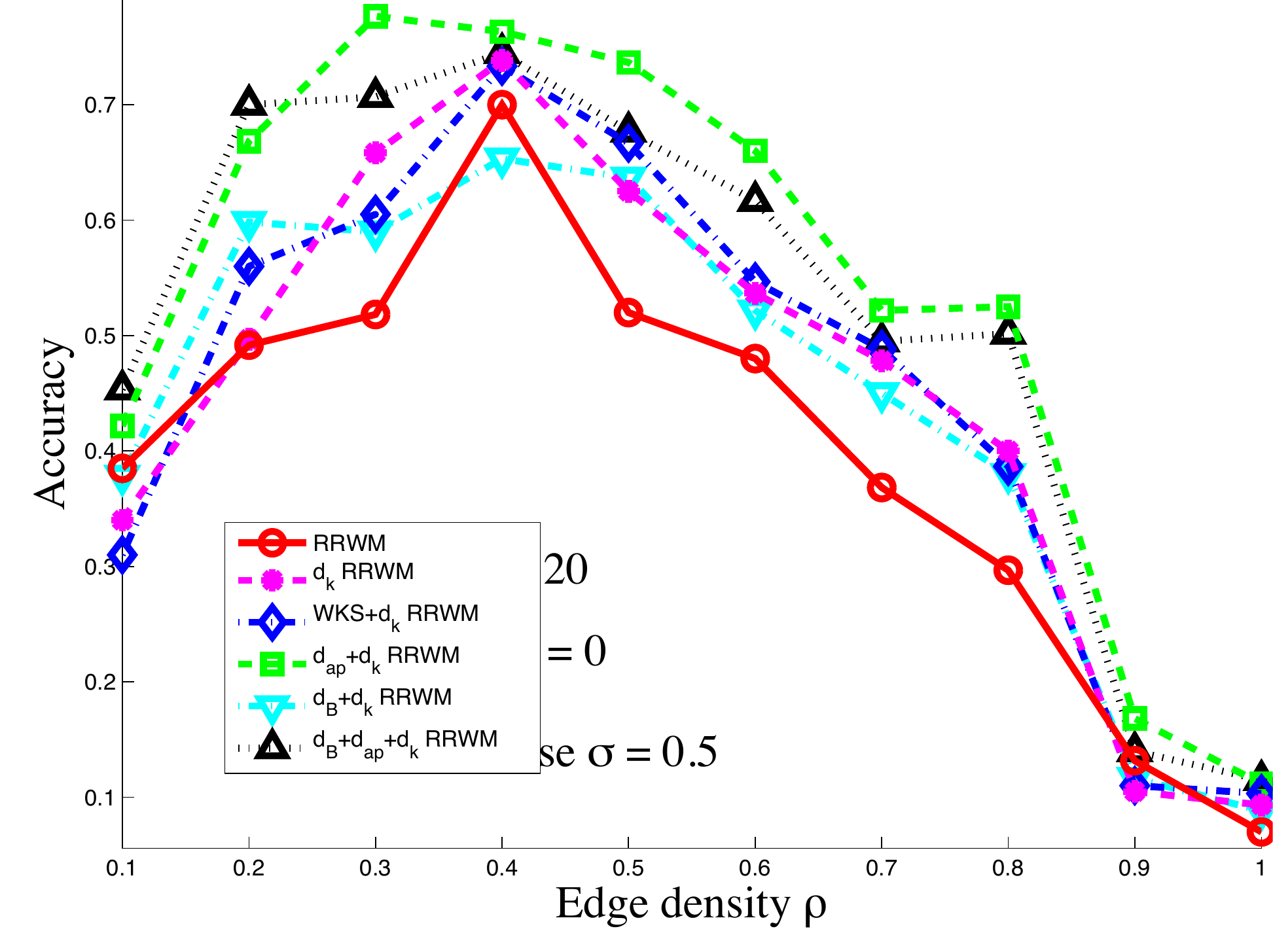}
\caption{}

\end{subfigure} \begin{subfigure}{.32\linewidth} \includegraphics[width=0.96\linewidth]{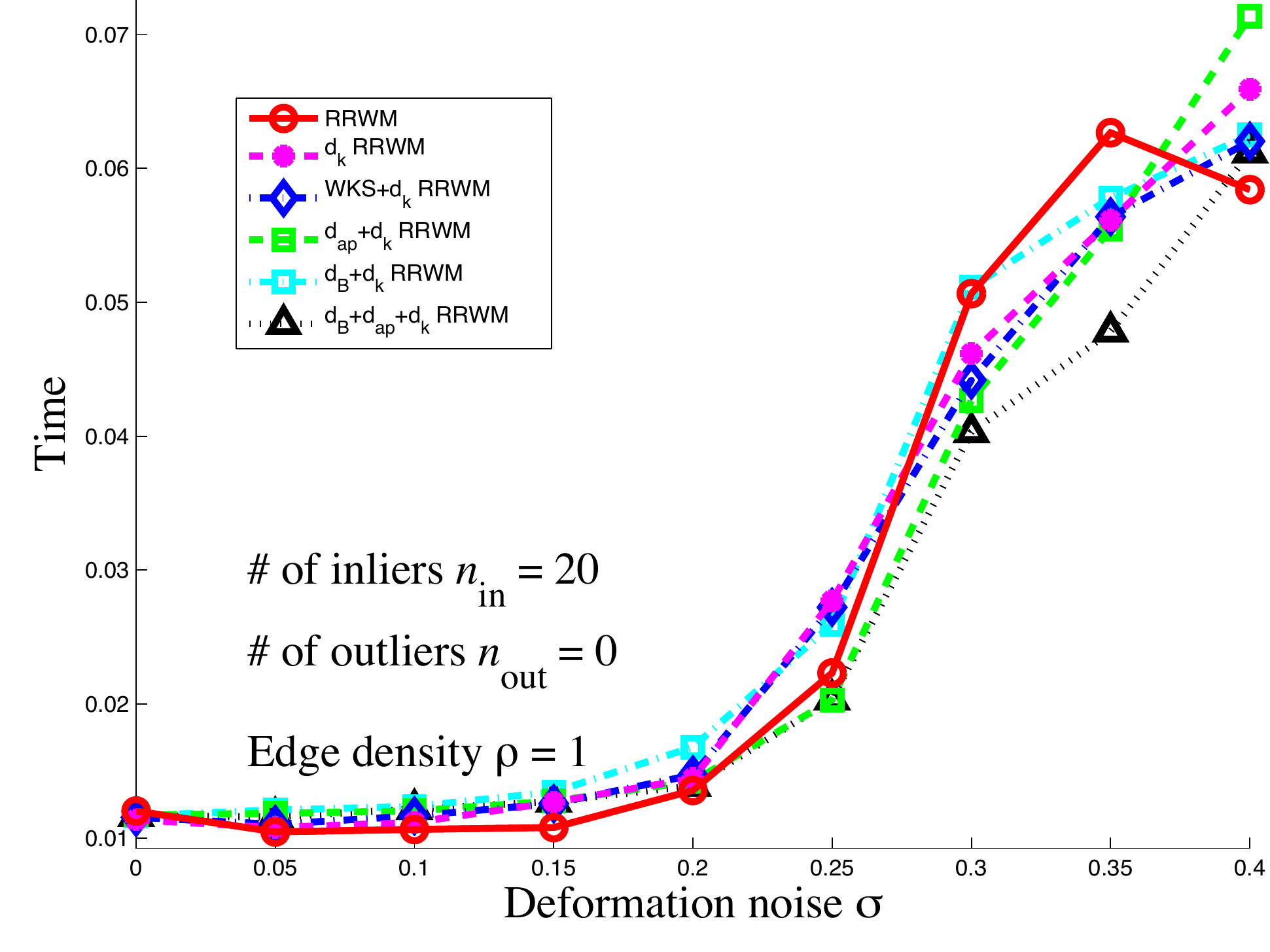}
\caption{}

\end{subfigure} \begin{subfigure}{.32\linewidth} \includegraphics[width=0.96\linewidth]{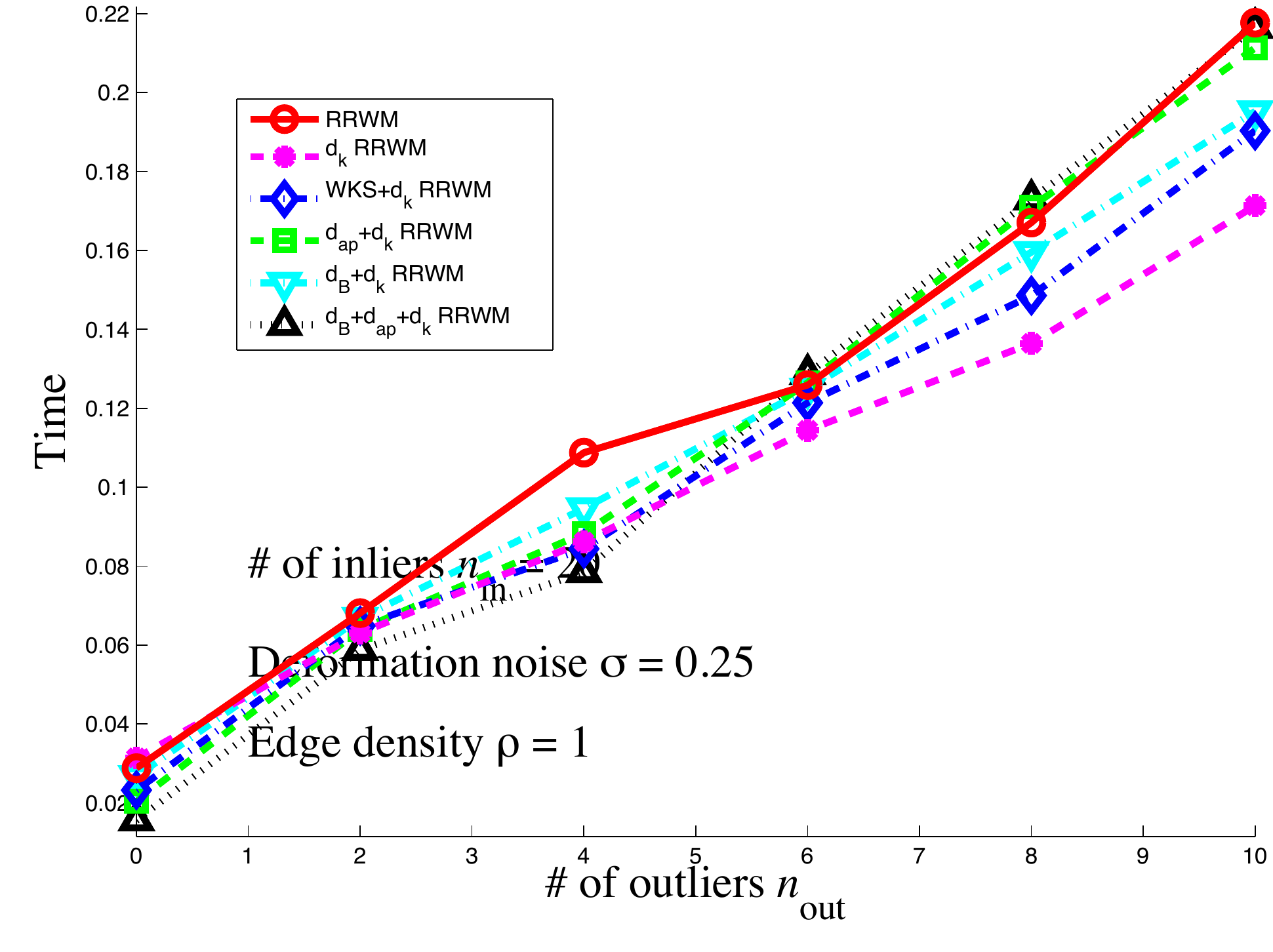}
\caption{}

\end{subfigure} \begin{subfigure}{.32\linewidth} \includegraphics[width=0.96\linewidth]{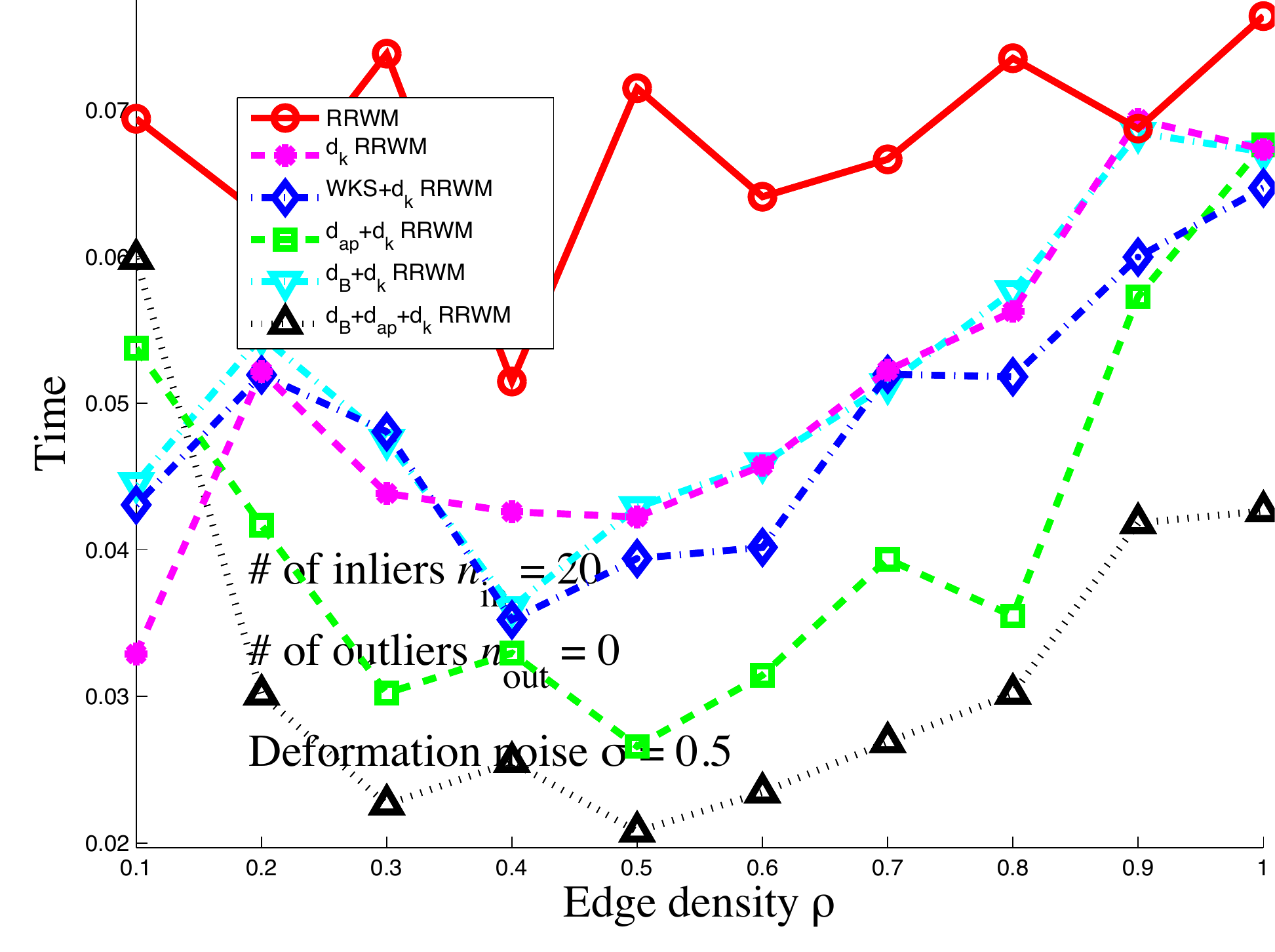}
\caption{}

\end{subfigure} \caption{Matching accuracy and computation time of IQP with different compatibility
functions.}

\label{fig:randiqp}
\end{figure*}

In this experiment, we test the matching performance for $W$ constructed
using i) only the adjacency matrix \cite{cho2010}, ii) only pairwise
heat kernel distance $d_{k}(i,j,a,b)$, iii) $d_{k}(i,j,a,b)$ with
WKS \cite{hu2012}, iv) $d_{k}(i,j,a,b)$ with $d_{\text{ap}}(i,a)$
($c_{B}=0$,$c_{\text{ap}}=1$), v) $d_{k}(i,j,a,b)$ with $d_{B}(i,a)$
($c_{B}=1$,$c_{\text{ap}}=0$), and vi) $d_{k}(i,j,a,b)$ with $d_{\text{ap}}(i,a)$
and $d_{B}(i,a)$ ($c_{B}=8$,$c_{\text{ap}}=3$), on three different
settings: 1) different levels of deformation noise $\sigma$; 2) different
numbers of outliers; 3) different edge densities $\rho$. Fig. \ref{fig:randiqp}
shows the average matching accuracy. In the experiment, the number
of anchor nodes $|U|=2$. The Red solid curve for RRWM is the baseline
approach using the adjacency matrices. From Fig. \ref{fig:randiqp}
(a), it can be seen that with the help of learned proximity matrix
$B$ and the term $d_{\text{ap}}(i,a)$, the matching results are
more robust to noise. In Fig. \ref{fig:randiqp} (c), the matching
accuracy is much improved at different edge densities for a relatively
large deformation noise ($\sigma=0.5$). Not only the matching accuracy
is improved, their corresponding computational time is also decreased
as shown in Fig. \ref{fig:randiqp} (f).

\subsection{CMU Hotel Sequence}

\begin{figure}[ht]
\begin{subfigure}{.96\linewidth} \centering \includegraphics[width=0.96\linewidth]{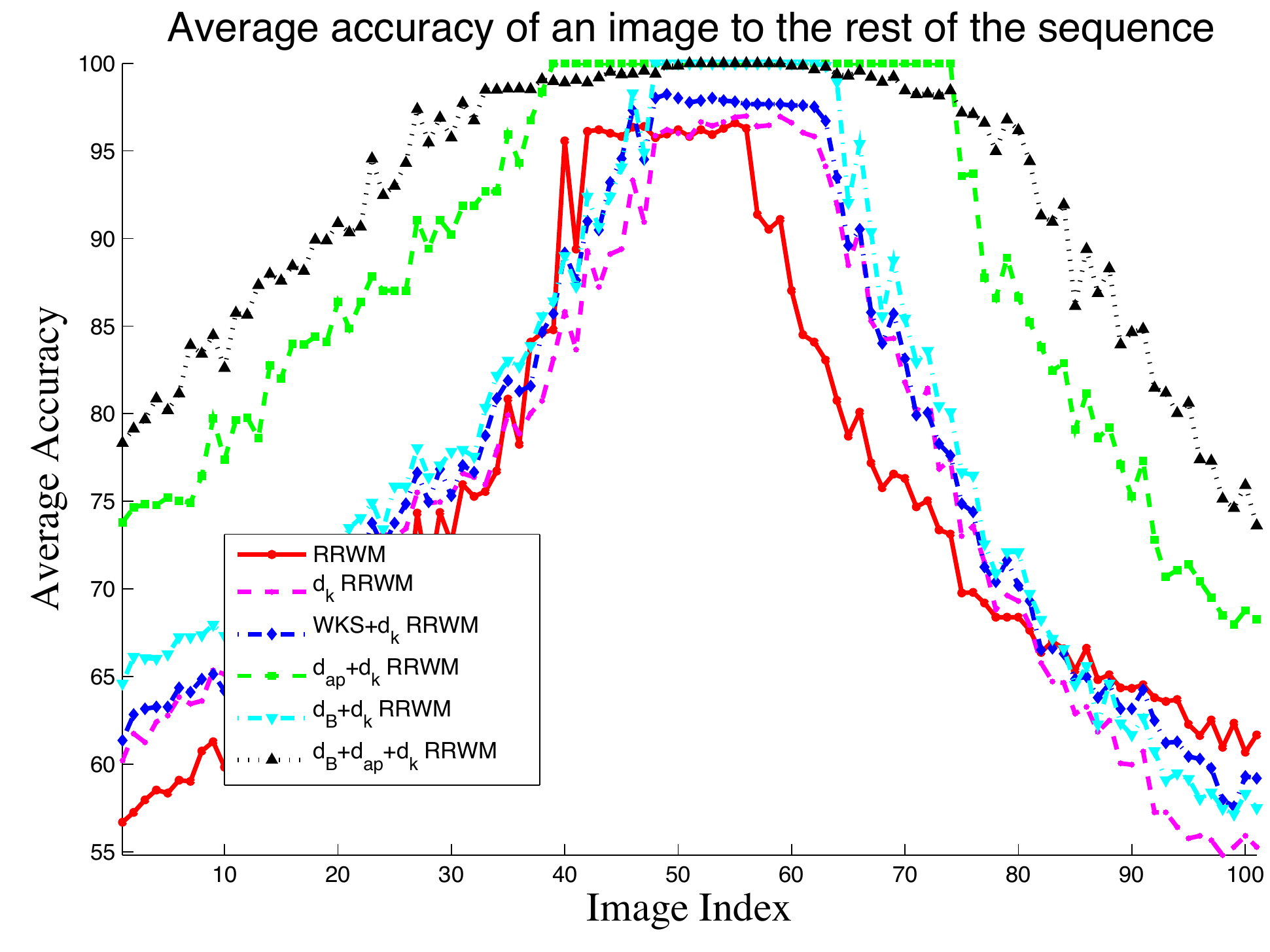}
\caption{Matching performance}

\end{subfigure}\\
 \centering \begin{subfigure}{.8\linewidth} \centering \includegraphics[width=0.96\linewidth]{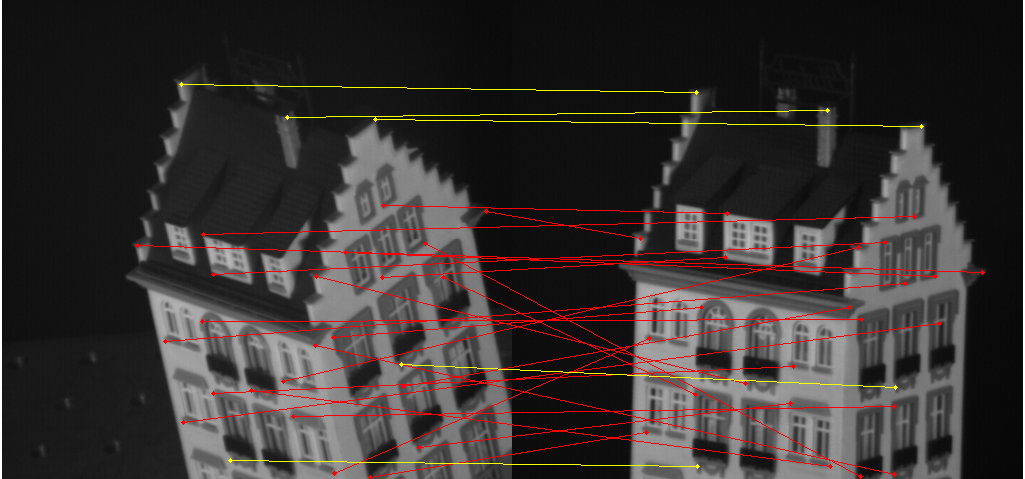}
\caption{RRWM}

\end{subfigure}\\
 \centering \begin{subfigure}{.8\linewidth} \centering \includegraphics[width=0.96\linewidth]{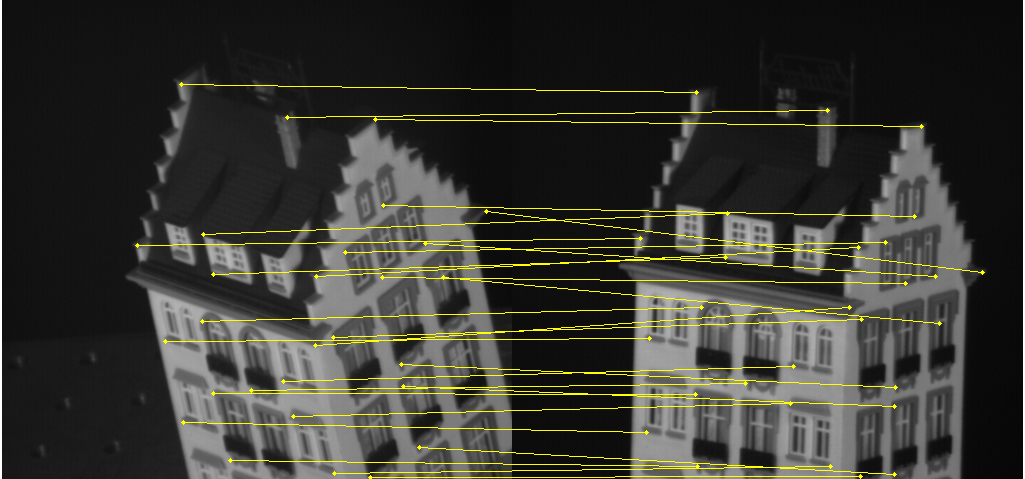}
\caption{$d_{B}+d_{\text{ap}}+d_{k}$ RRWM}

\end{subfigure} \caption{Matching on Hotel sequence. Yellow lines depict the correct matches,
while red lines show the wrong matches.}

\label{fig:houseplot}
\end{figure}

In this experiment, we test our descriptors on the CMU Hotel sequence,
which is widely used in performance evaluation of graph matching algorithms
as a wide baseline dataset. It consists of 101 frames, and there are
30 feature points labeled consistently across all frames. We build
fully connected graphs purely based on the geometry of the feature
points, taking the Euclidean distance as the weights between pair
of feature points. Affinity matrix $W$ were set up similarly as in
Section \ref{ssect:randgraph}. $|U|=2$ nodes were randomly selected
as the anchor nodes. We compute the average matching accuracy of each
frame to the rest of frames in the sequence. Fig. \ref{fig:houseplot}
(a) showed the performance of the matching. As can be seen the matching
performance was improved when heat kernel is used in lieu of the adjacency
matrix, because the noise tolerance property of the former smoothes out the effect
of deformation noise. With the add-on effect of proximity matrix $B$
and $d_{\text{ap}}(i,a)$, furthermore, the matching performance was
much improved. Fig. \ref{fig:houseplot} (b,c) showes an example of
the matching between the 20th and the 90th frame of the sequence (yellow
lines are correct matches and red lines are wrong matches). Matching
gives all correct matches, hence is only shown once.

In the second part of this experiment, we select to test the influence of the number of anchor nodes on the average matching rate. We intentionally drop off the term $d_{\text{ap}}$ to reduce the side effect, i.e. the matching scheme will be based on $d_{B}+d_{k}$. We increase the number of anchor nodes and compare the matching performance in otherwise the same setting.

\begin{figure}[ht]
\centering \includegraphics[width=0.96\linewidth]{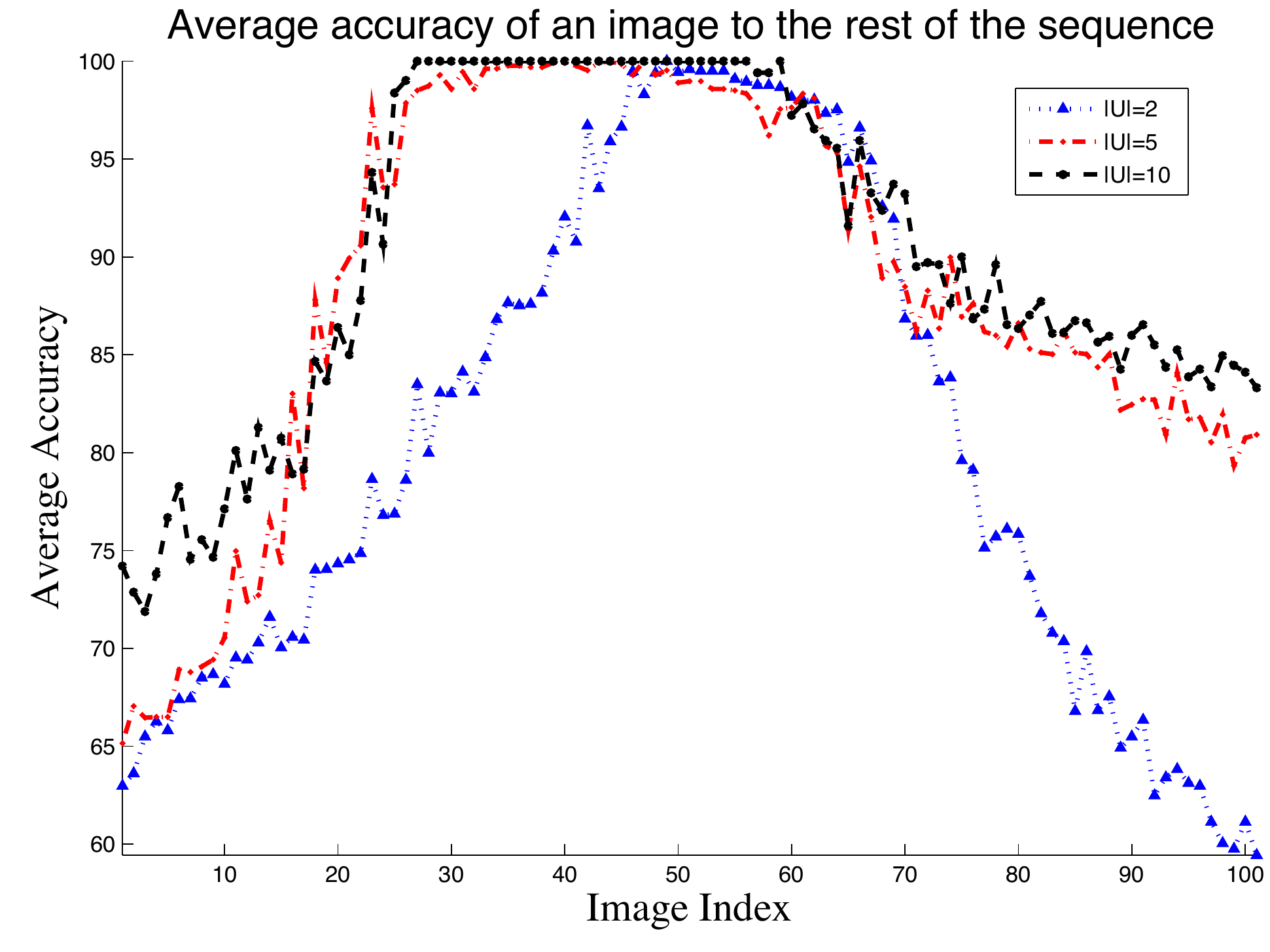}
\caption{Matching on Hotel sequence with different number of anchor nodes. (Matching scheme is $d_{B}+d_{k}$.)}
\label{fig:hotelnap}
\end{figure}

As shown in Fig. \ref{fig:hotelnap}, with the increase of the number of anchor nodes, the overall matching performance increased. However, the marginal performance gain seems to have a drop-off with the increase of the number of anchor nodes, since the matching rate gap between $|U|=10$ and $|U|=5$ is much smaller than that between $|U|=5$ and $|U|=2$.

\subsection{Pose House Sequence}

\begin{figure}[ht]
\centering \begin{subfigure}{.96\linewidth} \centering \includegraphics[width=0.96\linewidth]{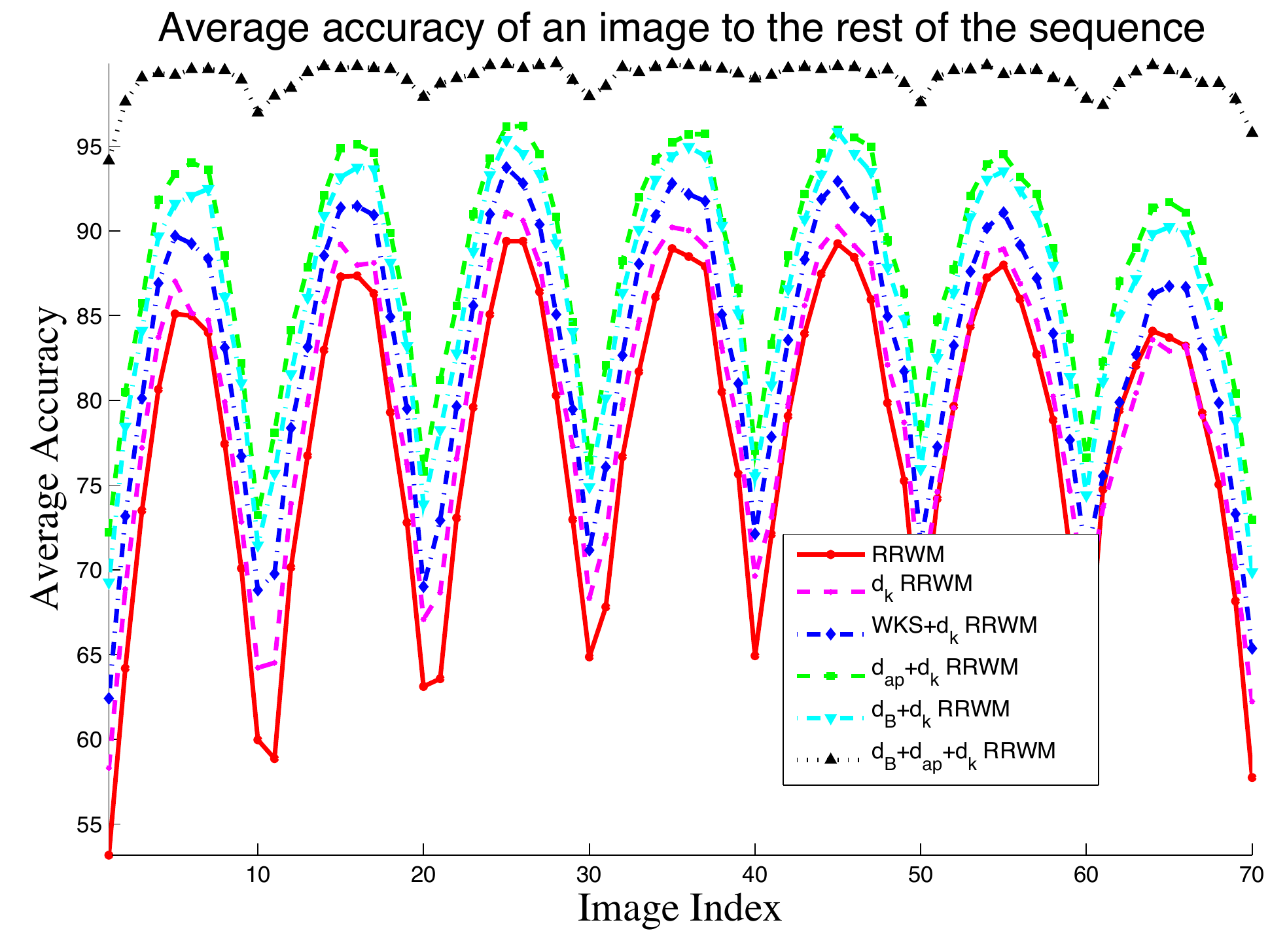}
\caption{Matching performance}

\end{subfigure}\\
 \begin{subfigure}{.8\linewidth} \centering \includegraphics[width=0.96\linewidth]{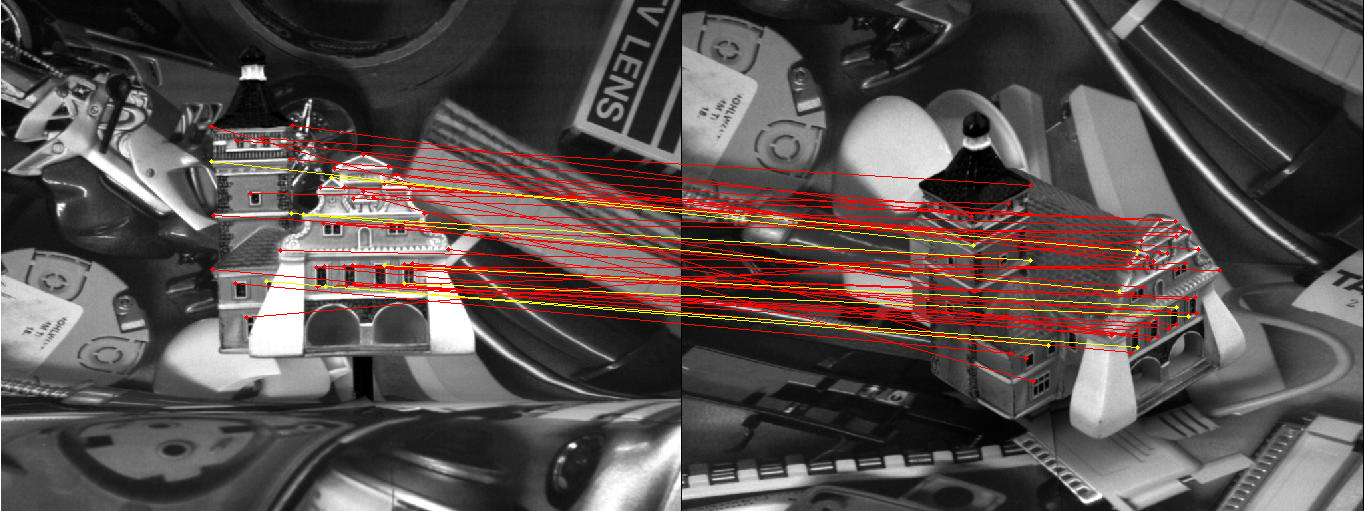}
\caption{RRWM}

\end{subfigure}\\
 \centering \begin{subfigure}{.8\linewidth} \centering \includegraphics[width=0.96\linewidth]{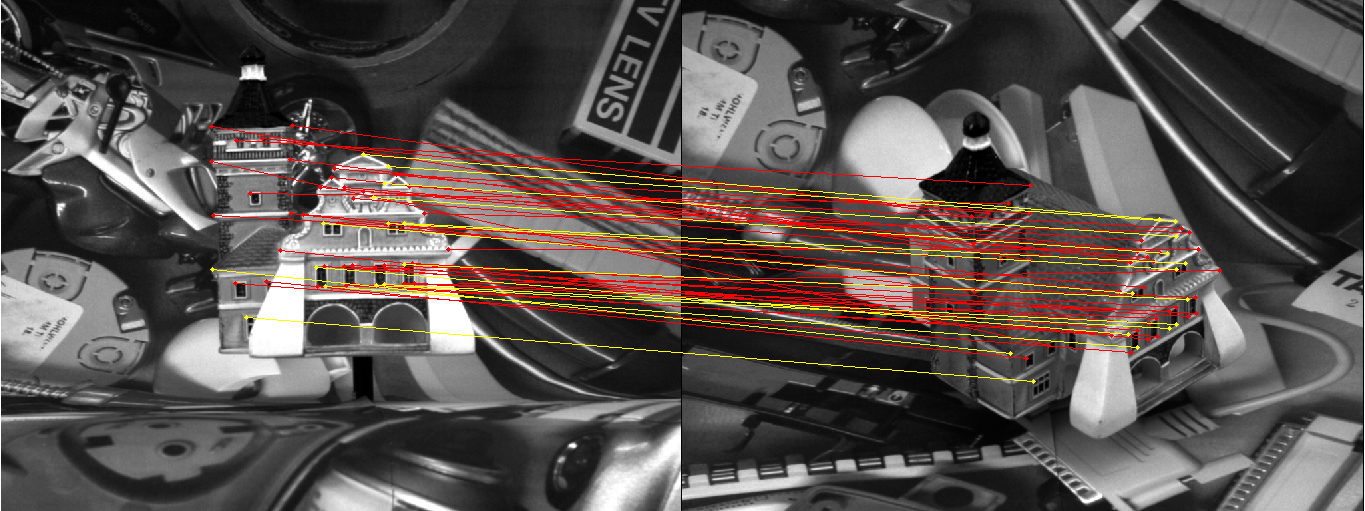}
\caption{$d_{\text{ap}}+d_{k}$ RRWM}

\end{subfigure}\\
 \centering \begin{subfigure}{.8\linewidth} \centering \includegraphics[width=0.96\linewidth]{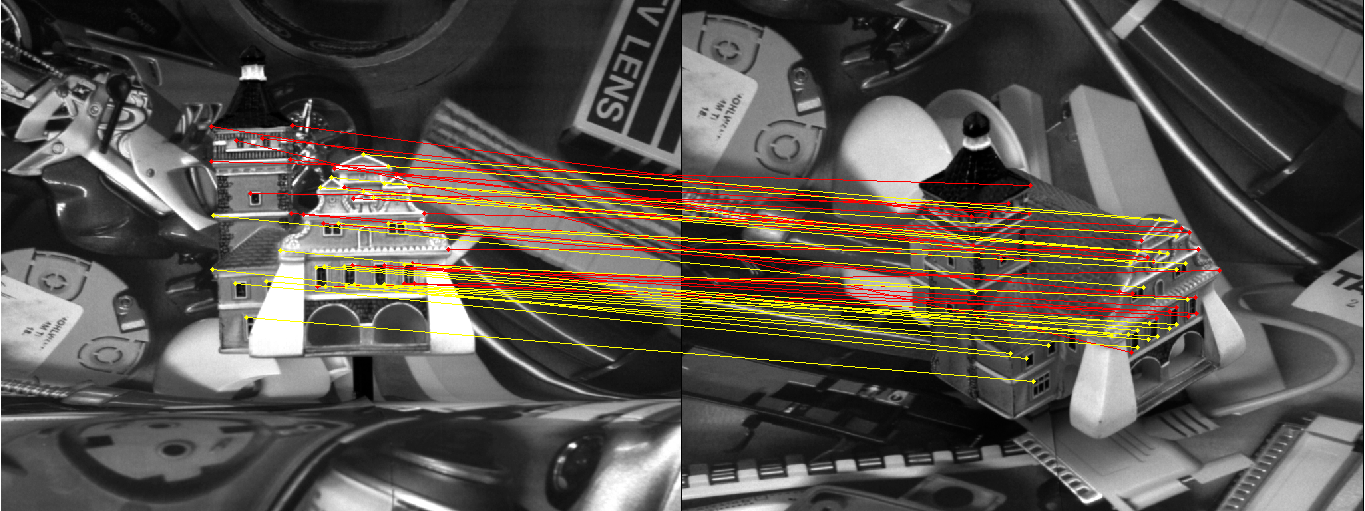}
\caption{$d_{B}+d_{\text{ap}}+d_{k}$ RRWM}

\end{subfigure} \caption{Matching on pose house sequence. Yellow lines depict the correct matches,
while red lines show the wrong matches.}

\label{fig:posehouse}
\end{figure}

In this experiment, we test our descriptor on the pose house sequence
used in \cite{mcauley2010}. The dataset consists of 70 frames with
51 labeled feature points across the sequence. The house undergoes
large pan and tilt angle change ($0-45^{\circ}$ for pan angle and$0-$$30^{\circ}$
for tilt). The compatibility matrix $W$ is built the same way as
in previous experiments. $|U|=2$ nodes were randomly selected as
the anchor nodes. Fig. \ref{fig:posehouse} showed the matching results.
In Fig. \ref{fig:posehouse} (a), each lump from left to right represent
a tilt angle from 0 to $30^{\circ}$ with a $5^{\circ}$ step and
within a lump is the pan angle change from left to right for 0 to
$45^{\circ}$ with a $5^{\circ}$step. It can been seen that extreme
pan angles gave poor results while mid range pan angles yield matches with
much higher accuracy --- while matching accuracy was not much influenced
by the difference in tilt angles. With the addition of our learned proximity
matrix $B$, the gap between different pan angles decreases and the
overall matching accuracy is superior to others. Fig. \ref{fig:posehouse}
(b,c) shows the matching results and the first and last frame of the
sequence, which represent the largest pan and tilt angles. Even in
this extreme case, it can be seen that our $d_{B}+d_{\text{ap}}+d_{k}$
matching gives useful results, and is much better than the adjacency
matrix based matching.

\section{Conclusion}

\label{sect:conclusion} In this paper, we have considered the problem
of graph matching where some correspondences are known. We have designed
a learning algorithm which uses the anchor correspondences as training
samples. The matching problem is set up as an IQP, where we use the
learned proximity and heat kernel distance to anchor nodes as first
order compatibility, and pairwise heat kernel distance difference
as a second order compatibility. With a very small number of anchor
nodes ($|U|=2$ in all of the experiments) we have obtained superior performance
as compared to the state-of-the-art techniques based on adjacency
matrices.

{\bf Acknowledgment}: The authors would like to acknowledge NSF grants IIS 1016324, CCF 1161480, DMS 1228304, AFOSR FA9550-12-1-0372, the Max Planck Center for Visual Computing and Communications, and a Google research award.

{\small{} \bibliographystyle{ieee}
\bibliography{hks}

\begin{thebibliography}{10}\itemsep=-1pt

\bibitem{Aubry2011}
M.~Aubry, U.~Schlickewei, and D.~Cremers.
\newblock The wave kernel signature: A quantum mechanical approach to shape
  analysis.
\newblock In {\em ICCV Workshop 4DMOD}, 2011.

\bibitem{biyi2007}
T.~Biyikoglu, J.~Leydold, and P.~F. Stadler.
\newblock {\em Laplacian eigenvectors of graphs: Perron-Frobenius and
  Faber-Krahn Type Theorems}.
\newblock Springer, 2007.

\bibitem{caetano2009}
T.~Caetano, J.~McAuley, C.~L., Q.~Le, and A.~Smola.
\newblock Learning graph matching.
\newblock {\em IEEE Trans. PAMI}, 31(6):1048--1058, 2009.

\bibitem{cho2010}
M.~Cho, J.~Lee, and K.~M. Lee.
\newblock Reweighted random walks for graph matching.
\newblock In {\em Proceedings of the 11th European conference on Computer
  vision: Part V}, ECCV'10, pages 492--505, Berlin, Heidelberg, 2010.
  Springer-Verlag.

\bibitem{cour2006}
T.~Cour, P.~Srinivasan, and J.~Shi.
\newblock Balanced graph matching.
\newblock In {\em NIPS'06}, pages 313--320, 2006.

\bibitem{emms2009}
D.~Emms, R.~C. Wilson, and E.~R. Hancock.
\newblock Graph matching using the interference of continuous-time quantum
  walks.
\newblock {\em Pattern Recogn.}, 42(5):985--1002, May 2009.

\bibitem{eshera1984}
M.~Eshera and K.~Fu.
\newblock A graph distance measure for image analysis.
\newblock {\em IEEE Trans. Syst. Man Cybern.}, page 398Ð408, 1984.

\bibitem{gold1996}
S.~Gold and A.~Rangarajan.
\newblock A graduated assignment algorithm for graph matching.
\newblock {\em IEEE Trans. Patt. Anal. Mach. Intell.}, 18, 1996.

\bibitem{gori2005}
M.~Gori, M.~Maggini, and L.~Sarti.
\newblock Exact and approximate graph matching using random walks.
\newblock {\em IEEE Trans. Pattern Anal. Mach. Intell.}, 27(7):1100--1111, July
  2005.

\bibitem{HAMMOND2011}
D.~K. Hammond, P.~Vandergheynst, and R.~Gribonval.
\newblock {Wavelets on graphs via spectral graph theory}.
\newblock {\em Applied and Computational Harmonic Analysis}, 30(2):129--150,
  Mar. 2011.

\bibitem{hu2012}
N.~{Hu} and L.~{Guibas}.
\newblock {Spectral Descriptors for Graph Matching}.
\newblock {\em ArXiv e-prints}, Apr. 2013.
\newblock arXiv:1304.1572[cs.CV].

\bibitem{jouili2009}
S.~Jouili and S.~Tabbone.
\newblock Graph matching based on node signatures.
\newblock In {\em Proceedings of GbRPR '09}, pages 154--163, Berlin,
  Heidelberg, 2009. Springer-Verlag.

\bibitem{leord2005}
M.~Leordeanu and M.~Hebert.
\newblock A spectral technique for correspondence problems using pairwise
  constraints.
\newblock In {\em ICCV '05}, pages 1482--1489, Washington, DC, USA, 2005. IEEE
  Computer Society.

\bibitem{leord2009}
M.~Leordeanu, M.~Hebert, and R.~Sukthankar.
\newblock An integer projected fixed point method for graph matching and map
  inference.
\newblock In {\em NIPS}. Springer, December 2009.

\bibitem{Leordeanu2012}
M.~Leordeanu, R.~Sukthankar, and M.~Hebert.
\newblock Unsupervised learning for graph matching.
\newblock {\em Int. J. Comput. Vision}, 96(1):28--45, Jan. 2012.

\bibitem{Lubbecke2005}
M.~E. L\"{u}bbecke and J.~Desrosiers.
\newblock Selected topics in column generation.
\newblock {\em Oper. Res.}, 53(6):1007--1023, Nov. 2005.

\bibitem{luo2001}
B.~Luo and E.~R. Hancock.
\newblock Structural graph matching using the em algorithm and singular value
  decomposition.
\newblock {\em IEEE Trans. Pattern Anal. Mach. Intell.}, 23(10):1120--1136,
  Oct. 2001.

\bibitem{mcauley2010}
J.~J. McAuley, T.~de~Campos, and T.~S. Caetano.
\newblock Unified graph matching in euclidean spaces.
\newblock {\em IEEE Conf. CVPR}, pages 1871--1878, 2010.

\bibitem{qiu2006}
H.~Qiu and E.~R. Hancock.
\newblock Graph matching and clustering using spectral partitions.
\newblock {\em Pattern Recogn.}, 39(1):22--34, Jan. 2006.

\bibitem{robles-kelly2002}
A.~Robles-Kelly and E.~R. Hancock.
\newblock String edit distance, random walks and graph matching.
\newblock In {\em Proceedings of the Joint IAPR International Workshop on
  Structural, Syntactic, and Statistical Pattern Recognition}, pages 104--112,
  London, UK, UK, 2002. Springer-Verlag.

\bibitem{robles-kelly2005}
A.~Robles-Kelly and E.~R. Hancock.
\newblock Graph edit distance from spectral seriation.
\newblock {\em IEEE Trans. Pattern Anal. Mach. Intell.}, 27(3):365--378, Mar.
  2005.

\bibitem{schell2005}
C.~Schellewald and C.~Schn\"{o}rr.
\newblock Probabilistic subgraph matching based on convex relaxation.
\newblock In {\em EMMCVPR'05}, pages 171--186, Berlin, Heidelberg, 2005.
  Springer-Verlag.

\bibitem{shok2001}
A.~Shokoufandeh and S.~J. Dickinson.
\newblock A unified framework for indexing and matching hierarchical shape
  structures.
\newblock In {\em IWVF-4}, pages 67--84, London, UK, UK, 2001. Springer-Verlag.

\bibitem{sog-hks-09}
J.~Sun, M.~Ovsjanikov, and L.~Guibas.
\newblock A concise and provably informative multi-scale signature based on
  heat diffusion.
\newblock In {\em Eurographics Symposium on Geometry Processing (SGP)}, 2009.

\bibitem{umeyama88}
S.~Umeyama.
\newblock An eigendecomposition approach to weighted graph matching problems.
\newblock {\em IEEE Trans. Pattern Anal. Mach. Intell.}, 10(5):695--703, 1988.

\bibitem{vanwyk2004}
B.~J. van Wyk and M.~A. van Wyk.
\newblock A pocs-based graph matching algorithm.
\newblock {\em IEEE Trans. Pattern Anal. Mach. Intell.}, 26(11):1526--1530,
  Nov. 2004.

\bibitem{Wong2006}
P.~C. Wong, H.~Foote, G.~Chin, P.~Mackey, and K.~Perrine.
\newblock {Graph signatures for visual analytics.}
\newblock {\em IEEE trans. on visualization and computer graphics},
  12(6):1399--413, 2006.

\bibitem{Xing2002}
E.~P. Xing, A.~Y. Ng, M.~I. Jordan, and S.~J. Russell.
\newblock Distance metric learning with application to clustering with
  side-information.
\newblock In {\em NIPS}, pages 505--512, 2002.

\bibitem{zas2009}
M.~Zaslavskiy, F.~Bach, and J.-P. Vert.
\newblock A path following algorithm for the graph matching problem.
\newblock {\em IEEE Transactions on PAMI}, 31(12):2227--2242, 2009.

\bibitem{zass2008}
R.~Zass and A.~Shashua.
\newblock Probabilistic graph and hypergraph matching.
\newblock {\em CVPR}, 2008.

\bibitem{zhao2007}
G.~Zhao, B.~Luo, J.~Tang, and J.~Ma.
\newblock Using eigen-decomposition method for weighted graph matching.
\newblock {\em ICIC'07}, pages 1283--1294, 2007.

\end{thebibliography}
 }
\end{document}